\def\year{2020}\relax
\documentclass[letterpaper]{article} 
\usepackage{aaai20}  
\usepackage{times}  
\usepackage{helvet} 
\usepackage{courier}  
\usepackage[hyphens]{url}  
\usepackage{graphicx} 
\usepackage{epstopdf}
\usepackage{bm}
\usepackage{amsmath}
\usepackage{amsfonts}
\usepackage{booktabs}
\usepackage{multirow}
\usepackage{array}
\usepackage{subfigure}
\newcommand{\citet}[1]{\citeauthor{#1}~\shortcite{#1}}
\title{Improving Neural Relation Extraction with Positive and Unlabeled Learning}
\author{Zhengqiu He$^\clubsuit$ \and Wenliang Chen $^\clubsuit$\thanks{The corresponding author is Wenliang Chen.}
\and  Yuyi Wang$^\heartsuit$ \\
       {\bf \Large Wei Zhang$^\spadesuit$ \and Guanchun Wang$^\diamondsuit$
\and Min Zhang$^\clubsuit$} \\
$^\clubsuit$Institute of Artificial Intelligence, School of Computer Science and Technology, Soochow
University, China\\
$^\spadesuit$Alibaba Group, $^\heartsuit$ETH Zurich, $^\diamondsuit$Laiye Network technology co.LTD\\
$^\clubsuit$zqhe@stu.suda.edu.cn, \{wlchen, minzhang\}@suda.edu.cn\\
$^\spadesuit$lantu.zw@alibaba-inc.com, $^\heartsuit$yuwang@ethz.ch, $^\diamondsuit$arvid@laiye.com\\
}

\begin{document}

\maketitle

\begin{abstract}
We present a novel approach to improve the performance of distant supervision relation extraction with Positive and Unlabeled (PU) Learning. This approach first applies reinforcement learning to decide whether a sentence is positive to a given relation, and then positive and unlabeled bags are constructed. In contrast to most previous studies, which mainly use selected positive instances only, we make full use of unlabeled instances and propose two new representations for positive and unlabeled bags. These two representations are then combined in an appropriate way to make bag-level prediction. Experimental results on a widely used real-world dataset demonstrate that this new approach indeed achieves significant and consistent improvements as compared to several competitive baselines.

\end{abstract}

\section{Introduction}
Relation extraction (RE) is to classify the relationship between entity pairs from plain text where the entity pairs are specified. Recently, supervised learning models have achieved lots of progress in the RE task \cite{riedel2013relation,Zeng2014Relation}. Similar to other Natural Language Processing (NLP) tasks, the supervised models for RE typically require a massive amount of training data, labeled by annotators. However, hiring annotators is costly and non-scalable, in terms of both time and money.

\begin{figure}[tb]
		\centering
\begin{minipage}{0.5\textwidth}
	\centering
	\includegraphics[width=0.92\textwidth]{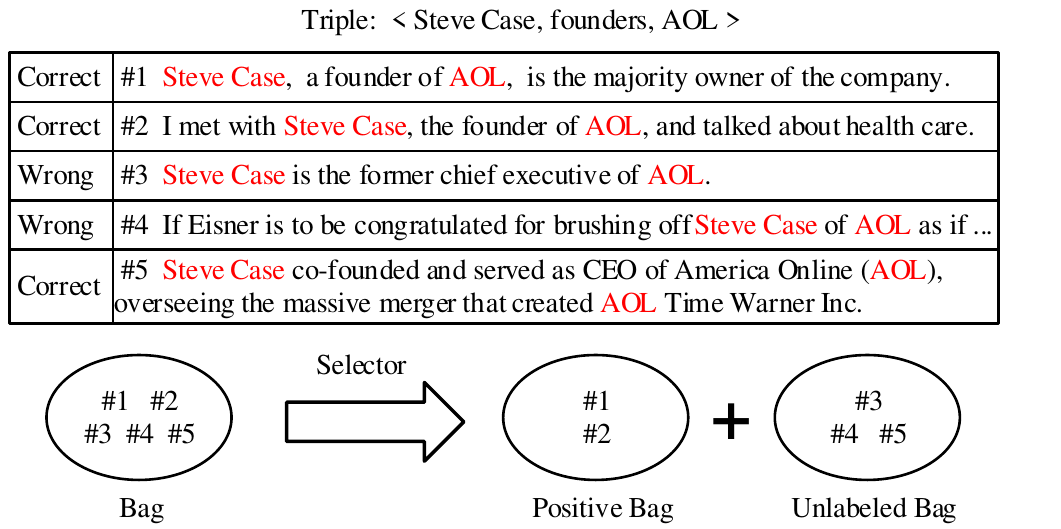}
\end{minipage}
\caption{A sentence bag generated through distant supervision, further split into two bags after selection}
\label{fig:example}
\end{figure}

As an alternative solution in practice,
distant supervision can automatically generate a training corpus \cite{mintz2009distant}.
The key idea of distant supervision is that given an entity pair $<\!\!e_h, e_t\!\!>$ and its corresponding relation $r_B$ from one knowledge base (KB) such as Freebase, we simply label all sentences containing the two entities by relation $r_B$. This solution has achieved a certain success and has been used in the state-of-the-art techniques for RE \cite{riedel2010modeling,hoffmann2011knowledge,surdeanu2012multi,min2013distant}. However, the training data generated by this solution inevitably has wrong labeling problem.
For example, as illustrated in Figure \ref{fig:example},
we have a triple including entity pair $<$\emph{Steve Case, AOL}$>$ and its relation ``\emph{founders}"\footnote{The relation ``/business/company/founders"  is defined in Freebase, and we shorten it to ``founders".}  and label five sentences by distant supervision.
Though the third sentence ``\emph{Steve Case is the former chief executive of AOL.}" does not convey relation ``\emph{founders}"
of ``\emph{Steve Case}" and ``\emph{AOL}", it is wrongly labeled as a positive instance for this relation under the assumption of distant supervision. Our preliminary experimental results show that this problem largely harms the performance of relation extraction  at \textbf{sentence-level}.


To address the above problem, previous studies turn to multi-instance learning (MIL) \cite{zeng2015distant,Lin2016Neural,ji2017distant}, where the training set consists of bags for given entity pairs.
Thus, the task of RE turns to predicting relations at  \textbf{bag-level}. However, because the wrong labeling problem is still remaining, each bag in multi-instance learning may contain many sentences mentioning the same entity pair but belonging to different relations. These noisy sentences may hinder the performance of the MIL-based relation extraction models.

To alleviate the impact of noisy sentences, recent studies \cite{feng2018rl,Qin2018robust} propose to select positive sentences from noisy bags. As a selection solution, approaches based on reinforcement learning (RL) have been proposed to train RL agents to decide whether a sentence is positive.
Thus the sentences are grouped into a set of positive instances and a set of unlabeled instances, as shown in Figure \ref{fig:example}\footnote{Please note sentence \#5 is positive in Figure \ref{fig:example}, but it is grouped into the unlabeled bag by the sentence selector.}.
There have been three solutions to utilize the sets of positive and unlabeled instances,
which have achieved a certain success.
One solution ignores the unlabeled sentences and only keeps the positive sentences in MIL to perform \textbf{bag-level} prediction at final relation extractor \cite{zeng2018rl}.
The second solution discards the unlabeled sentences and utilizes the positive sentences to perform \textbf{sentence-level} prediction at final relation extractor \cite{feng2018rl}.
The third solution treats the unlabeled sentences as negative instances and utilizes both positive and negative instances to perform \textbf{sentence-level} prediction at final relation extractor \cite{Qin2018robust}.
However, we argue that using only the selected positive instances and ignoring the other sentences when making relation prediction at \textbf{bag-level} is suboptimal, and
treating the unlabeled sentences as negative instances might not be appropriate since they might contain positive instances (for example, sentence \#5 in Figure \ref{fig:example}).

\begin{figure}[tb]
	\centering
	\begin{minipage}{0.5\textwidth}
		\centering
		\includegraphics[width=0.92\textwidth]{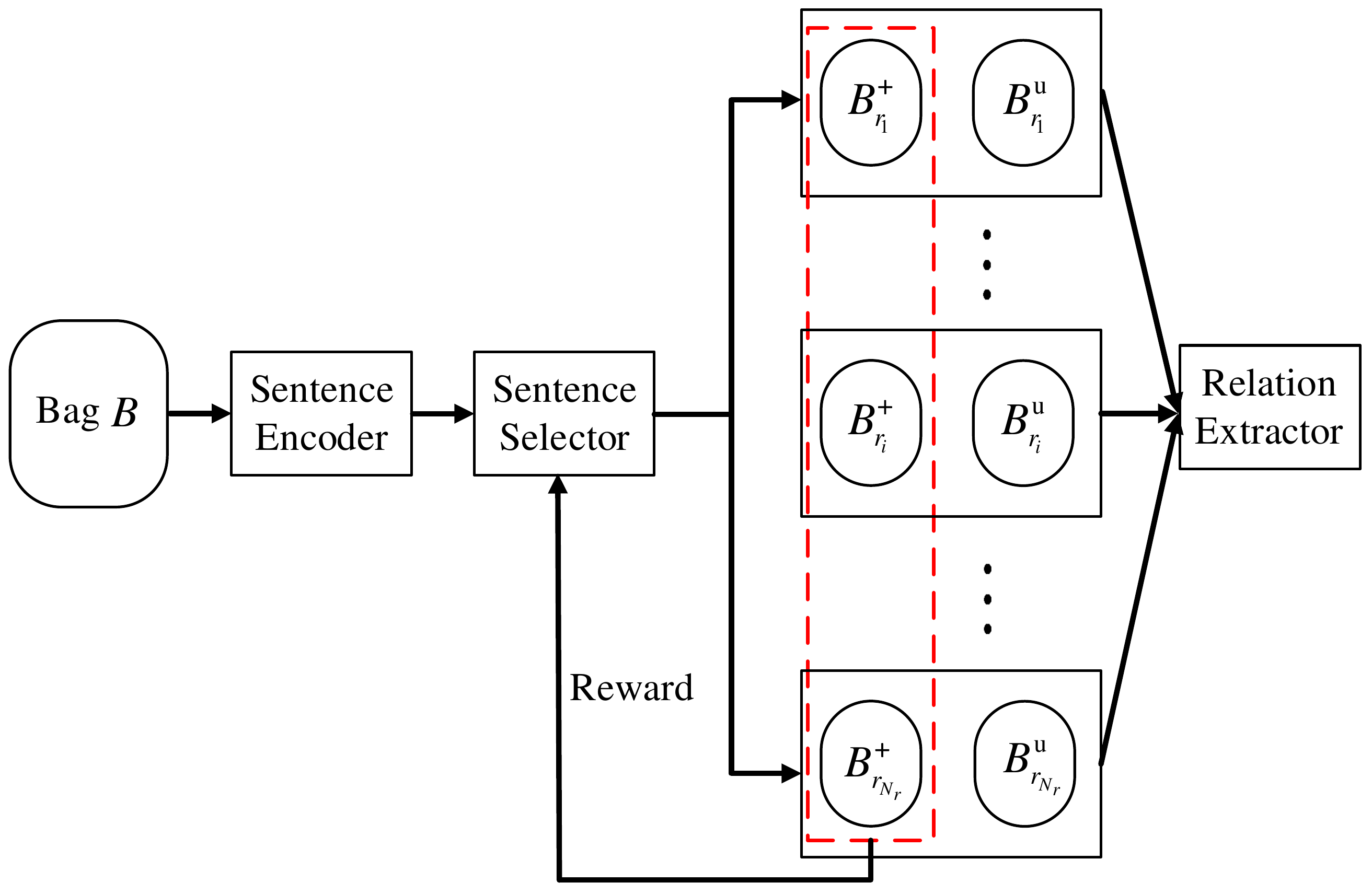}
	\end{minipage}
	\caption{Infrastructure of our proposed model}\label{arch}
\end{figure}

Ideally, we would wish to combine the merits of all solutions in our final relation extractor, that is using both positive and unlabeled instances in MIL for \textbf{bag-level} prediction.
In this paper, we first train an RL agent to automatically select the positive sentences for each relation. As shown in Figure \ref{fig:example}, for relation ``founders", the agent selects the first two sentences as a new positive bag, and the last three sentences are in an unlabeled bag.
Therefore, our task is converted into a situation similar to a positive and unlabeled (PU) learning problem \cite{10.1007/11564096_24,10.1007/978-3-662-44845-8_5}.
Classical PU learning methods, either selecting negative data points from the unlabeled collection, or simply assuming all unlabeled data points are negative \cite{10.1007/11564096_24,10.1007/978-3-662-44845-8_5}, are not suitable for our task as we mention above. Thus we propose a new method to exploit both positive and unlabeled data.
That is, we first build two representations for positive and unlabeled sentences respectively, and then the two representations are combined into one as a \textbf{bag-level} representation which helps with making \textbf{bag-level} prediction.

In summary, we make the following contributions:
\begin{itemize}
	\item  In our RL-based instance selector, we design a new state representation which considers sentence embeddings, relation embeddings, and the embeddings of selected positive instances. To our best knowledge, it is the first time that the relation embedding is used in the state representation. Experimental results show that the relation embedding can significantly improve our system, as shown in Figures \ref{fig:cnn_baseline} and \ref{fig:pcnn_baseline}.

\item In order to exploit unlabeled sentences properly,
	we propose a model to build representations for positive and unlabeled sentences in a
	bag respectively and then they are combined into one representation. Totally, we have three kinds of representations for relation extractor. The ablation study in experiments shows that all the representations make positive contributions to our system, as shown in Figures \ref{fig:cnn_ab} and \ref{fig:pcnn_ab}.

\end{itemize}

We design experiments to demonstrate the effectiveness of our system. The results on a widely used dataset indicate that our proposed model significantly outperforms the systems using RL-based instance selector in the previous studies.

\section{Our Approach}


In this section, we describe our approach in detail. Due to the distant supervision assumption, the sentences in the bag of an entity pair may not belong to the predefined relation type. The basic idea is that, we identify the relevance of a relation and every sentence in a bag, and output positive and unlabeled bags for the relation.
This selection procedure can be considered as a sequential decision-making process which can be modeled by reinforcement learning.
Based on both positive and unlabeled bags, we build our relation extractor.

Our framework consists of three key components as shown in Figure \ref{arch}: (1) A sentence encoder, which encodes the sentences into distributed representations.
(2) An instance selector based on the RL agent, which decides whether a sentence is positive for each query relation. The agent takes the distributed representation of sentence and relation as input. (3) A relation extractor, which is trained on the positive and unlabeled bags generated by the sentence selector.


\subsection{Sentence Encoder}

The sentence encoder transforms a sentence into its distributed representation by a (P)CNN \cite{Zeng2014Relation,zeng2015distant}, which has shown the state-of-the-art performance in this task.

\subsubsection{Input Representation}


Given a sentence
$q$,
and $q_i$ is the $i$-th word in the sentence, the input is a matrix composed of $n$ vectors
$\bm{w} = [\bm{w_1}, ..., \bm{w_n}]$, where $\bm{w_i}$ corresponds to $q_i$ and consists of its word embedding and  position embedding. Following previous work \cite{zeng2015distant,Lin2016Neural}, we employ the skip-gram method \cite{mikolov2013b} to pretrain the word embedding, which will be fine-tuned afterwards.
Position embedding is first successfully applied to relation extraction task by \citet{Zeng2014Relation}, which specifies the relative distances of a word with respect to two target entities. In this way, $\bm{w_i} \in \mathbb{R}^d $, and $d = d^a + 2 \times d^b$, where $d^a$ and $d^b$ are the dimensions of word embedding and position embedding respectively.

\subsubsection{Convolution, Max-pooling}

After encoding the input words, a convolution layer and a max-pooling operation are applied to reconstruct the original input $\bm{w}$ by learning sentence features from a small window of words at a time while preserving word order information.
We use $K$ convolution filters with the same window size $l$.
Formally, the $j$-th element of the output vector $\bm{c} \in \mathbb{R}^{K}$ is
\begin{equation}
\bm{c}[j] = \mathop{\mathrm{max}}\limits_{i}(\bm{W}_j\bm{w}_{i:i+l-1})
\end{equation}
where $\bm{W_j}$ is the weight matrix.
%
%

Further, \citet{zeng2015distant} adopt piecewise max pooling (PCNN) in relation extraction, which is a variation of CNN. Suppose the positions of the two entities are $p_1$ and $p_2$ respectively. Then, each convolution output vector is divided into three segments:
\begin{equation}
[0:p_1-1]/[p_1:p_2]/[p_2+1:n-l]
\end{equation}

Then the max-pooling operation is applied to each segment.
Finally, we apply a non-linear transformation such as $\mathrm{tanh}$ on the output vector
to obtain the sentence embedding.
For each sentence $q$ in bag $B$, we get sentence embedding $\bm{x_q}$ from the sentence encoder.

\subsection{Sentence Selector}

The sentence selector is an RL agent which takes a query relation and a sentence as inputs
and decides whether the sentence is positive to the relation.
Here we suppose that the inputs are query relation $r$ and sentence $q$ (with its
embedding $\bm{x_q}$) from bag $B$.
As for bag $B$, after selection we obtain positive bag $B^\textit{+}_r$
and unlabeled bag $B^\textit{u}_r$.
We first define state representation, Q-network policy function, and reward function.
Then we describe how to train the RL agent.


\subsubsection{State Representation}

In our state representation, we first consider the information from the sentence as \citet{feng2018rl} do.
To model our task as a Markov decision process (MDP) problem \cite{fang2017learning},
we add the information from the early states into the representation of the current state.
In addition, we consider the information of the relation since it is one of the key factors of selection.
Thus, the state representation includes:
(1) The sentence embedding $\bm{x_q}$.
(2) The relation embedding $\bm{r_e}$ of relation $r$, which is initialized randomly and tuned during training.
(3) The embedding $\bm{x_{pos}}$, which is the average vector of the positive sentences
selected so far in the current bag.
Therefore, the state representation $S_{q,r}$ is $[\bm{x_q},\bm{r_e},\bm{x_{pos}}]$. Compared with the state presentation used in \citet{feng2018rl}, ours considers the information of relation embedding.

\subsubsection{Q-network}
The agent takes an action to decide whether sentence $q$ is positive to
query relation $r$.
The action is taken by
the policy function $A_{q,r} \! =\! \mathrm{argmax}~Q(S_{q,r},A)$,
where $Q$ is the state-action table \cite{watkins1992q}, and $A$ is a discrete set of actions. 
Following \citet{mnih2015human}, we employ a two-layer neural networks as the policy function. In our case, the action space includes two actions: positive (+1) and unlabeled (-1).

\subsubsection{Reward}
We regard the selection of a bag as an episode,
and a reward is calculated after all positive sentences are selected.
Thus we only have a delayed reward without immediate rewards in bag $B$.
We provide the delayed reward as feedback to the agent.
We set the reward to be the conditional probability that the selected positive bag $B^\textit{+}_r$ belongs to the query relation. For query $r$ and Bag $B$, the reward function is defined as
\begin{equation}
R(r,B) = P(r|B_r^\textit{+})
\end{equation}
%
%
%
The conditional probability $P(r|B^\textit{+}_r)$ is computed as
\begin{equation}\label{eq:pos_prob}
P(r|B_r^\textit{+}) = \frac{\mathrm{exp}(\mathbf{o}_r^\textit{+}[r])}
{\sum_{1 \le k \le N_r} {\mathrm{exp}(\mathbf{o}_r^\textit{+}[r_k])}}
\end{equation}
where 
$N_r$ is the total number of relations, and $\mathbf{o}_r^\textit{+}$ 
corresponds to the score vector associated to all relation types:
\begin{equation}\label{eq:pos_rep}
\mathbf{o}_r^\textit{+} = \bm{W}\textbf{emb}_{B_r^\textit{+}}+\bm{b}
\end{equation}
where $\bm{W}$ is a weight matrix, $\bm{b}$ is a bias vector,
and $\textbf{emb}_{B^\textit{+}_r}$  is the corresponding embedding of bag $B^\textit{+}_r$.

\subsubsection{Training RL Agent}
We aim to maximize the expected total reward for all bags, and train the agent using Q-learning \cite{watkins1992q}, a standard reinforcement learning algorithm that 
learns policies for an agent interacting with its environment. 
Each episode during training corresponds to
a query relation and a bag, where the agent selects the positive sentences of the relation from the bag.

Let $Q(S,A)$ be the measure of long-term cumulative reward obtained by taking action $A$ from state $S$. We consider the function approximation parameterized by $\theta^Q$, which are optimized by minimizing the loss:
\begin{equation}
L(\theta^Q) = \mathbb{E}_{S,A}[(V_t - Q(S,A|\theta^Q)^2]
\end{equation}
where $V_t$ is estimated using the Bellman equation \cite{sutton1998reinforcement}:
\begin{equation}
V_t = \mathbb{E}_{S'}[R(S,A) + \gamma \mathop{\mathrm{max}}\limits_{A'}Q(S',A'|\theta^Q)]
\end{equation}
where $R(S,A)$ is the immediate reward, and $\gamma$ is the discount factor for the value of future rewards.
As mentioned above, we don't have the immediate reward.
And the order of sentences in a bag should not influence the result.
Thus, the immediate reward is set to 0, and $\gamma$ is set to 1.
As for $r$ and $B$, we have
\begin{equation}
V_t = R(r,B)
\end{equation}

We optimize the loss using stochastic gradient descent (SGD) and experience replay, with random mini-batch of past experience sampled for training.

\subsection{Relation Extractor}


After the process of the sentence selector, we build our relation extractor based on both positive and unlabeled bags. The reason we involve the unlabeled instances is to increase the robustness of the method in noisy environment of distant supervision.

We first define three types of representations from different views and then define loss functions based on the representations. Finally, we jointly train the relation extractor and the sentence selector.

\subsubsection{Three Representations}
We generate three types of representations:
(1) POS-based representation, which represents the selected positive bag.
(2) UNL-based representation, which represents the unlabeled bag.
(3) PU-combined representation, which is a combination of the above two representations.
For all three representations, we project to vectors in $N_r$-dimension space.

\textbf{POS-based Representation} $\mathbf{o}_r^\textit{+}$: Based on the positive bag $B^{+}_r$ and relation $r$, we obtain the POS-based representation $\mathbf{o}_r^\textit{+}$ by Equation (\ref{eq:pos_rep}) which is also used for the reward function of the sentence selector. We share the parameters of $\mathbf{o}_r^\textit{+}$ between the relation extractor and the sentence selector.

\textbf{UNL-based Representation} $\mathbf{o}_r^\textit{u}$: Based on the unlabeled bag $B^{u}_r$ and relation $r$, the UNL-based representation $\mathbf{o}_r^\textit{u}$ is computed as
\begin{equation}
\mathbf{o}_r^\textit{u} = \bm{W}^{u}\textbf{emb}_{B^\textit{u}_r}+\bm{b}^{u}
\end{equation}
where $\bm{W}^{u}$ is a parameter matrix, and $\bm{b}^{u}$ is a bias vector.

\textbf{PU-combined Representation} $\bm{o}_r$: This type of representation is a combination of the above two representations, which is inspired by the idea of the feature combination in traditional models.
The PU-combined representation $\bm{o}_r$ is computed as
\begin{equation}
\bm{o}_r = \alpha\bm{o}_r^\textit{+} + (1-\alpha)\bm{o}_r^\textit{u}
\end{equation}
where $\alpha\in(0,1)$ is a weight for combination.

\subsubsection{Three Loss Functions}
We define three loss functions based on the representations respectively.
Here we suppose that training data $\mathcal{D} = \{(B_1,r_{B_1}), ..., (B_M, r_{B_M})\}$ consists of $M$ sentence bags and their corresponding relation labels assigned by distant supervision.

As for POS-based representation, we use Equation (\ref{eq:pos_prob}) to calculate the probability $P(r|B_r^\textit{+})$. Then we define cross-entropy loss function as the training objective:
\begin{equation}\label{eq:agent-cross-entropy-loss}
L_{pos}(\mathcal{D}) = - \sum_{i=1}^{M}\mathrm{log}\,P(r_{B_i}|B_{ir_{B_i}}^\textit{+})
\end{equation}
where $B_{ir_{B_i}}^\textit{+}$ is the positive bag for $r_{B_i}$ from $B_i$.

As for UNL-based representation, the conditional probability $P(r|B^\textit{u}_r)$ for each $r$ is computed as
\begin{equation}
P(r|B^\textit{u}_r) = \frac{\mathrm{exp}(\mathbf{o}_r^\textit{u}[r])}
{\sum_{1 \le k \le N_r} {\mathrm{exp}(\mathbf{o}_r^\textit{u}[r_k])}}
\end{equation}

The loss function is defined as
\begin{equation}\label{eq:neg-cross-entropy-loss}
L_{unl}(\mathcal{D}) = - \sum_{i=1}^{M}\mathrm{log}\,(1-P(r_{B_i}|B_{ir_{B_i}}^\textit{u}))
\end{equation}
where $B_{ir_{B_i}}^\textit{u}$ is the unlabeled bag for $r_{B_i}$ from $B_i$.

As for PU-combined representation, we can compute the conditional probability $P(r|B)$ for bag B as
\begin{equation}\label{eq:bag-prob}
P(r|B) = \frac{\mathrm{exp}(\mathbf{o}_r[r])}
{\sum_{1 \le k \le N_r} {\mathrm{exp}(\mathbf{o}_r[r_k])}}
\end{equation}
Please note that for each query relation $r_i$, we will perform a sentence selection and calculate the probability $P(r_i|B)$. Thus the probability of one query relation is independent from the other types, and we try to optimize our model to enlarge the probability $P(r_B|B)$ of bag label $r_B$ for better prediction at bag-level. Therefore, we design a loss function to rank the given relation label higher than the other labels at bag-level:
\begin{equation}\label{eq:set-cross-entropy-loss}
\small
\begin{split}
L_{bag}\!(\mathcal{D}) \!\!=\!-\!\sum_{i=1}^{M}\!\Big(\!\mathrm{log}P(r_{B_i}\!|\!B_i\!) \!+\! \mathrm{log}(1\!-\!P(r_{B_i}^*\!|\!B_i))\!\Big)
\end{split}
\end{equation}
where $r_{B_i}$ is the given label of sentence bag $B_i$ of focused entity pair, $r_{B_i}^*$ is the relation type that have the highest conditional probability among the other relation labels.

Finally, we combine the three loss functions for final relation extraction:
\begin{equation}
L(\mathcal{D}) = L_{pos}(\mathcal{D}) + L_{unl}(\mathcal{D}) + \beta L_{bag}(\mathcal{D})
\end{equation}
where $\beta$ is a weighting factor.

\subsubsection{Training and Testing}
To solve the optimization problem, we adopt mini-batch SGD to minimize the objective function.
During learning, we jointly train the relation extractor and the RL agent until convergence.

In the testing phase,
for each input bag, we first use the sentence selector to output the positive and unlabeled bags for each relation. Then we use Equation (\ref{eq:bag-prob}) to
perform relation classification
based on the positive and unlabeled bags.

%
%

\section{Experiments}
In this section, we present our experimental results and detailed analysis.
\begin{figure*}[htb]
	\centering
	\begin{minipage}[t]{0.45\textwidth}
		\centering
		\includegraphics[width=0.95\textwidth]{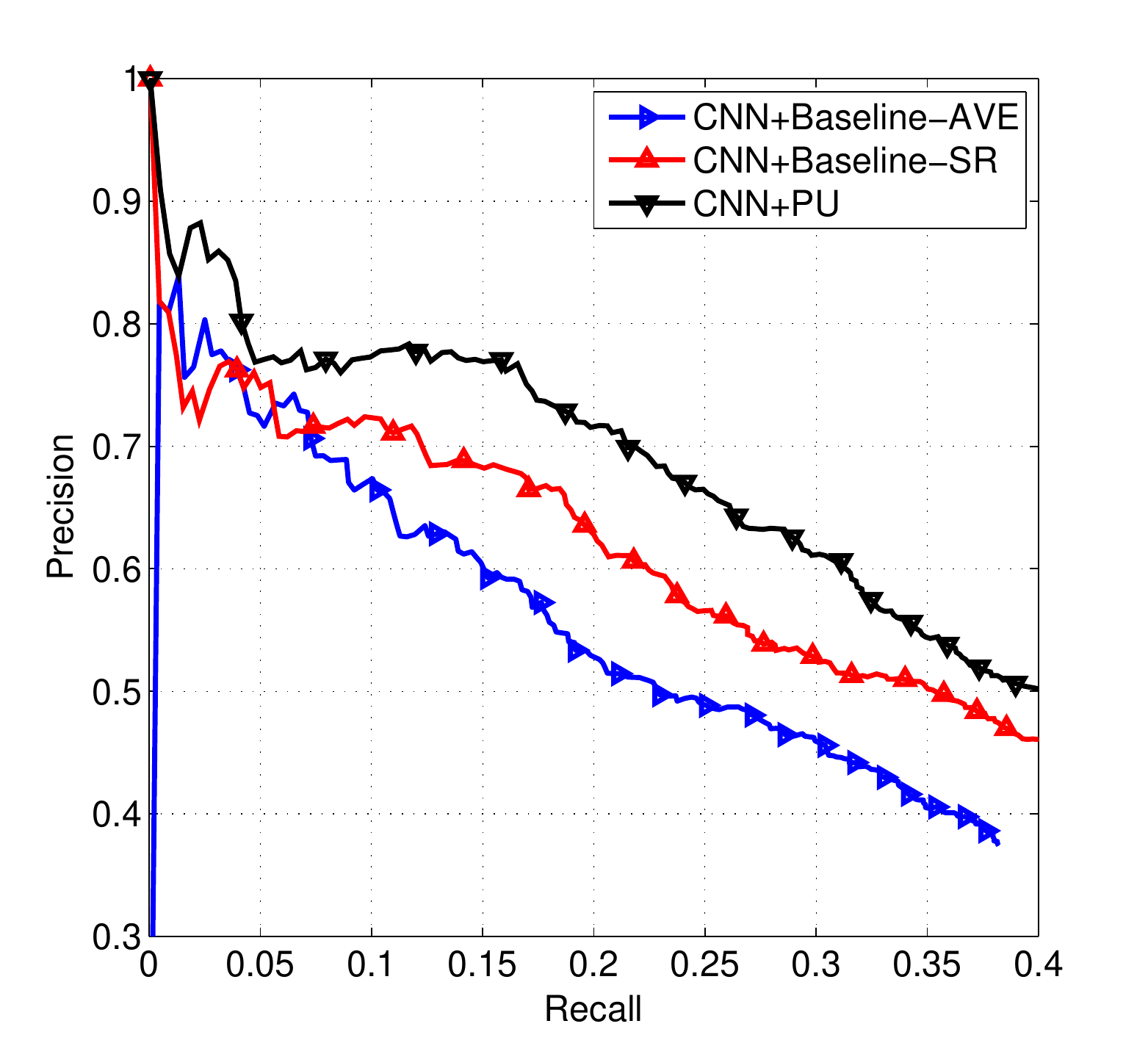}
		\caption{Comparison with baselines under CNN}\label{fig:cnn_baseline}
	\end{minipage}
	\begin{minipage}[t]{0.45\textwidth}
		\centering
		\includegraphics[width=0.95\textwidth]{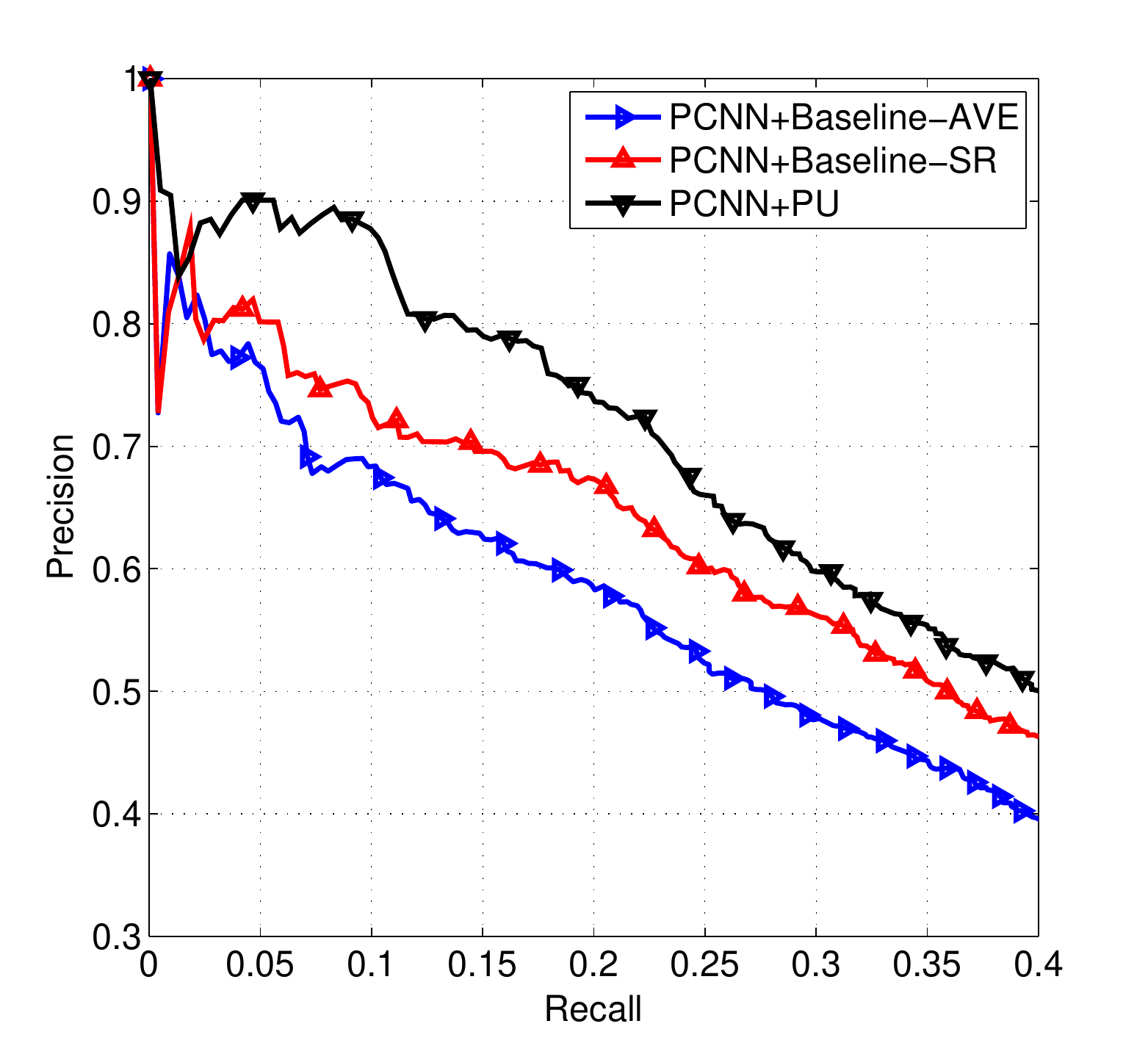}
		\caption{Comparison with baselines under PCNN}\label{fig:pcnn_baseline}
	\end{minipage}
\end{figure*}
\subsection{Datasets}

We evaluate our model on a widely used benchmark dataset developed by \citet{riedel2010modeling}, which has also been used in many recent studies \cite{Lin2016Neural,ji2017distant}.
The dataset contains 53 relation types, including a special relation ``NA" standing for no relation between the entity pair.
The training data contains 522,611 sentences, 281,270 entity pairs and 18,252 relational facts. The testing set contains 172,448 sentences, 96,678 entity pairs and 1,950 relational facts.

\subsection{Evaluation Metrics}
Following the practice of previous studies \cite{zeng2015distant,Lin2016Neural},
we evaluate our method in two ways, i.e., the \emph{held-out evaluation} and \emph{manual evaluation}.

The held-out evaluation does not consider the instances of the NA category,
and only compares the entity-relation tuples produced by the system on the test data against the existing Freebase entity-relation tuples. We report the precision-recall curves of the experiments.

The manual evaluation is performed to avoid the influence of the wrong labels resulting from distant supervision and the incompleteness on Freebase data. In the manual evaluation,
we manually check the newly discovered relation instances,
and reports the  precision of the top $N$ discovered relational facts with the highest probabilities.
We have two annotators which independently annotate the outputs. For divergent annotation results, discussions are held until agreement is reached.

\subsection{Hyperparameter Settings}
In the experiments, we use the same hyperparameters of sentence encoder as previous work \cite{zeng2015distant,Lin2016Neural},
and tune other hyperparameters via three-fold cross validation on training data.

After tuning, we choose the following settings in our experiments: the dimension of word embedding $d^a$ is set to 50, the dimension of position embedding $d^b$ is set to 5, the number of filters $K$ is set to 230, and the window size $l$ of filters is 3. The batch size is fixed to 50, the dropout probability is set to 0.5.  And the combination weight $\alpha$ is set to 0.7, $\beta$ is set to 0.1. When training, we apply Adam \cite{kingma2014adam} to optimize parameters, and the learning rate is set to 0.001.

\subsection{Main Results}

In this section, we compare our system (PU) with two baseline systems through held-out evaluation.
We select the CNN model proposed by \citet{Zeng2014Relation} and the PCNN model proposed by \citet{zeng2015distant} as our sentence encoders. The first baseline (Baseline-AVE) is a system without RL which represents the bag as the average vector of all the sentences in bags (described in \citet{Zeng2014Relation} and \citet{zeng2015distant}). The second one (Baseline-SR) is a system where we remove the relation embedding in the state representation in the sentence selector from our final system.

Figures \ref{fig:cnn_baseline} and \ref{fig:pcnn_baseline} show the results, where the higher the position of a curve is, the better the corresponding model is.
From the figures, we observe that for both CNN and PCNN, our system (PU) and Baseline-SR achieve better performance than Baseline-AVE. This fact indicates that, the RL-based model can effectively select positive sentences from bags and is beneficial to relation extraction. From the results of PU and Baseline-SR, we can also see that using relation embedding as the component of state representation is helpful.



\subsection{Comparison to Previous Approaches}
To evaluate our proposed method, we compare our model to several advanced systems through held-out evaluation, which fall into three categories:

\begin{itemize}
	
	\item Traditional discrete feature-based methods: 
	(1) \textbf{MultiR} \cite{hoffmann2011knowledge} is a probabilistic graphical model with multi-instance learning under the ``at-least-one" assumption. 
	(2) \textbf{MIML} \cite{surdeanu2012multi} is a graphical model with both multi-instance and multi-label learning.
	
	\item Attention-based methods:
	(1) \textbf{PCNN+ATT} \cite{Lin2016Neural} utilizes sentence-level attention to reduce the weights of noisy instances. (2) \textbf{APCNN+D} \cite{ji2017distant} uses external background information of entities via an attention layer to help relation classification.
	\item Reinforcement learning-based methods: (1) \textbf{CNN+RL} \cite{feng2018rl} proposes to use RL to select sentences in the training process, and predicts the relations at  sentence-level.
	(2) \textbf{PCNN+ATT+RL} \cite{Qin2018robust} proposes to redistribute the distant supervision data and then trains the PCNN+ATT model on the selected data.
	(3) \textbf{PE+REINF} \cite{zeng2018rl} regards the relation extractor as reinforcement learning agent, and follows the expressed-at-least-once assumption to predict the bag relation.
\end{itemize}

\begin{figure}
	\begin{minipage}[t]{0.45\textwidth}
		\centering
		\includegraphics[width=0.92\textwidth]{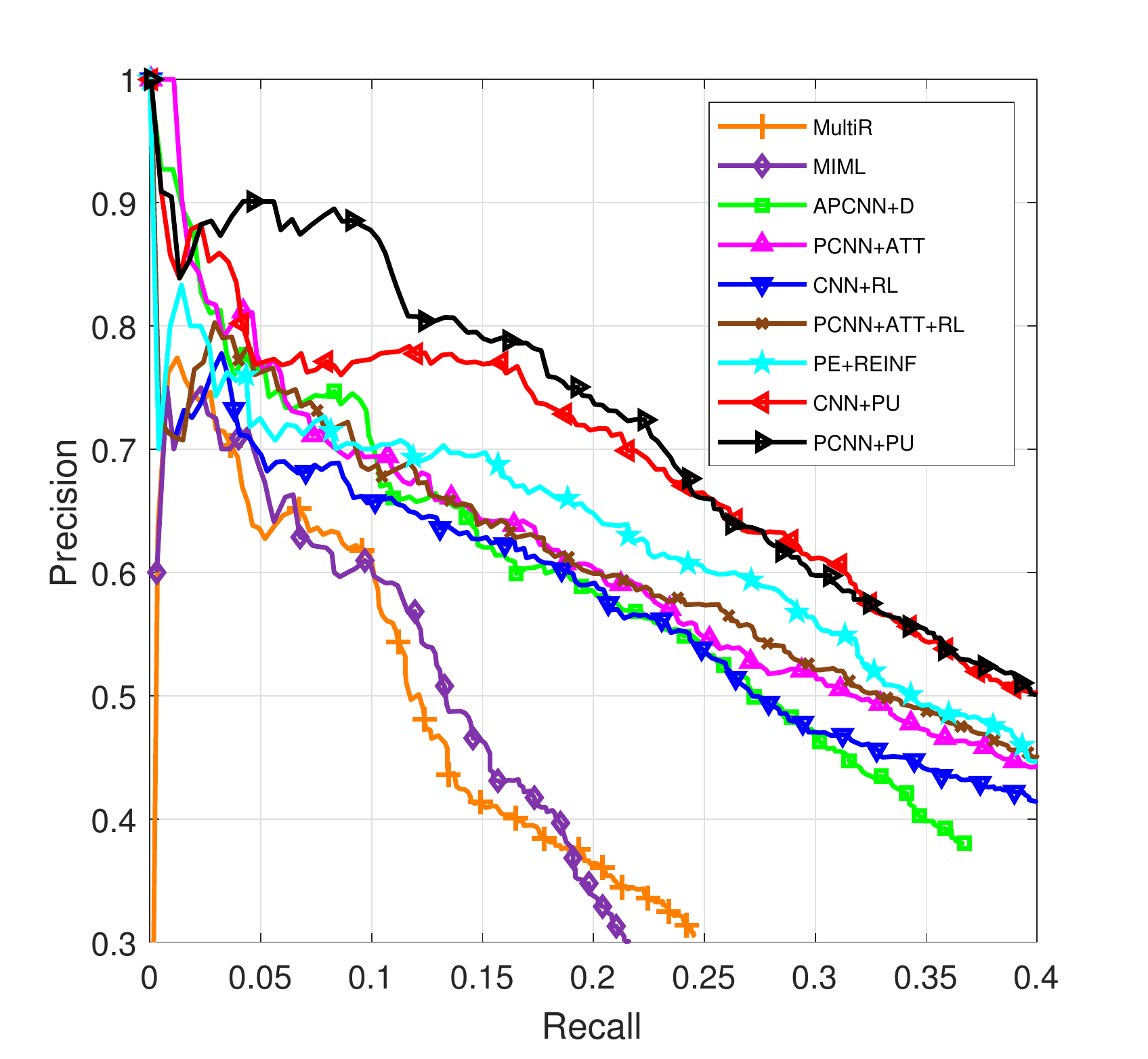}
		\caption{Comparison with previous results}\label{fig:proposed_baseline}
	\end{minipage}
\end{figure}

Figure \ref{fig:proposed_baseline} shows the precision/recall curves for these methods. We can observe that:
(1) Both MultiR and MIML underperform as compared to the Neural-based methods. It demonstrates that the human-designed feature cannot concisely express the semantic meaning of the sentences, and the inevitable error brought by NLP tools will hurt the performance of relation extraction.
(2) Both attention-based and reinforcement learning-based methods can obtain reasonable performance.
(3) Both CNN+PU and PCNN+PU consistently outperforms all other approaches by large margin. It indicates that the screening process of sentences and the joint training of positive and noise sentences are effective for distant supervision relation extraction.
(4) PCNN+PU performs a little better as compared to CNN+PU. It means that the piecewise max-pooling strategy is helpful for encoding sentences, and the model of our model can be further improved if we have a better sentence encoder.

\begin{table}[tb]
	\centering
	\small
	\begin{tabular}{p{58pt}<{\centering}|p{36pt}<{\centering}p{36pt}<{\centering}p{36pt}<{\centering}}
		\hline
		Accuracy & Top 100 & Top 200 & Top 500 \\
		\hline
		MultiR   & 0.83 &0.74 & 0.59  \\
		MIML     & 0.85 &0.75 & 0.61  \\
		APCNN+D  & 0.87 &0.83 & 0.74 \\
		PCNN+ATT & 0.86 &0.83 & 0.73 \\
		\hline
		PCNN+PU &\textbf{0.92} & \textbf{0.89} &\textbf{0.80}  \\
		\hline
	\end{tabular}
	\caption{Manual evaluation results}\label{manual}
\end{table}

\begin{table}[!tb]
	\centering
	\small
	\begin{tabular}{p{55pt}<{\centering}|p{42pt}<{\centering}|p{42pt}<{\centering}}
		\hline
		\multirow{1}{*}{Accuracy}
		&
		\multicolumn{1}{c|}{Top 100} &
		\multicolumn{1}{c}{Top 300} \\
		\cline{2-3}
		
		\hline
		CNN+PU&  0.96
		&  0.93  \\
		\hline
		PCNN+PU &  0.97
		& 0.95  \\
		\hline
	\end{tabular}
	
	\caption{Accuracy on relation extraction based on selected sentences}\label{table:manual_select}
\end{table}

\subsection{Manual Evaluation}
Due to existence of noises resulting from distant supervision and the incomplete nature of Freebase, a majority of the tuples that were produced with high confidence are false negatives and are actually true relation instances.
Thus, we can see that there is a sharp decline in the precision-recall curves in most models in Figure \ref{fig:proposed_baseline}.
Therefore, we perform a manual evaluation
to eliminate these problems, and report the $P$@100, $P$@200, $P$@500
for PCNN+PU and four baseline models, as shown in Table \ref{manual}.

We can see that PCNN+PU achieves
the best performance. Moreover, the precision is higher than in the held-out evaluation. It indicates that many of the false negatives that we predicted are actually true relational facts.

\subsection{Ablation Study}
In this section, we perform an ablation study of our system through held-out evaluation.  We want to see the contributions of the proposed representations in the model. There are three components of bag representations in our system. We compare the different settings via CNN/PCNN:
(1) \textbf{PU (Ours)} is our final system based on the proposed model.
(2) \textbf{PU-COMB} is a system where we remove the loss function of PU-combined representation from our final system (PU).
(3) \textbf{PU-COMB-UNL} is a system where we remove the loss function of UNL-based representation from PU-COMB. PU-COMB-UNL only uses the POS-based representation with RL-based sentence selector.

\begin{figure*}[htb]
    \centering
	\begin{minipage}[t]{0.45\textwidth}
		\centering
		\includegraphics[width=0.95\textwidth]{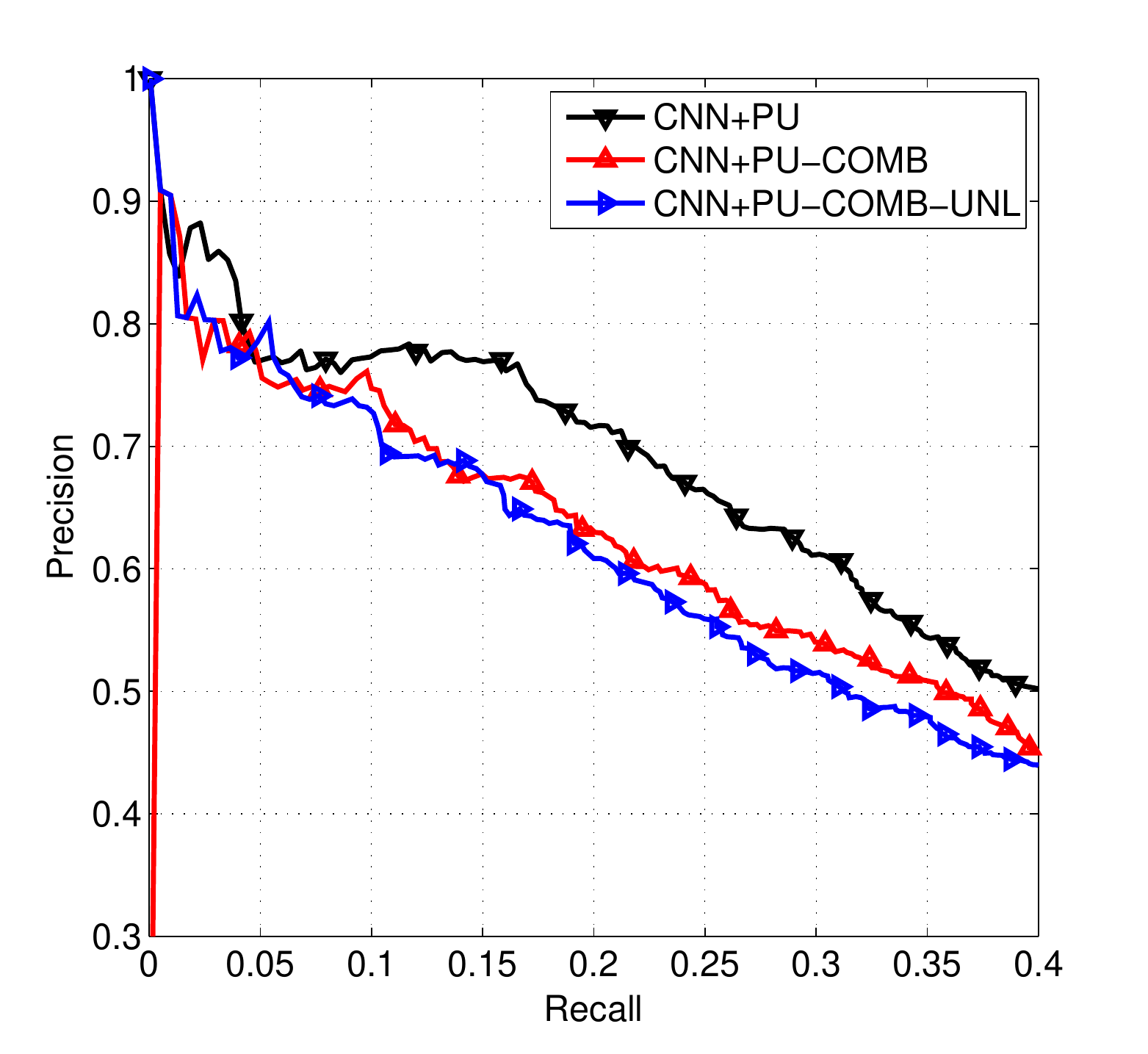}
		\caption{Ablation Study under CNN}\label{fig:cnn_ab}
	\end{minipage}
	\begin{minipage}[t]{0.45\textwidth}
		\centering
		\includegraphics[width=0.95\textwidth]{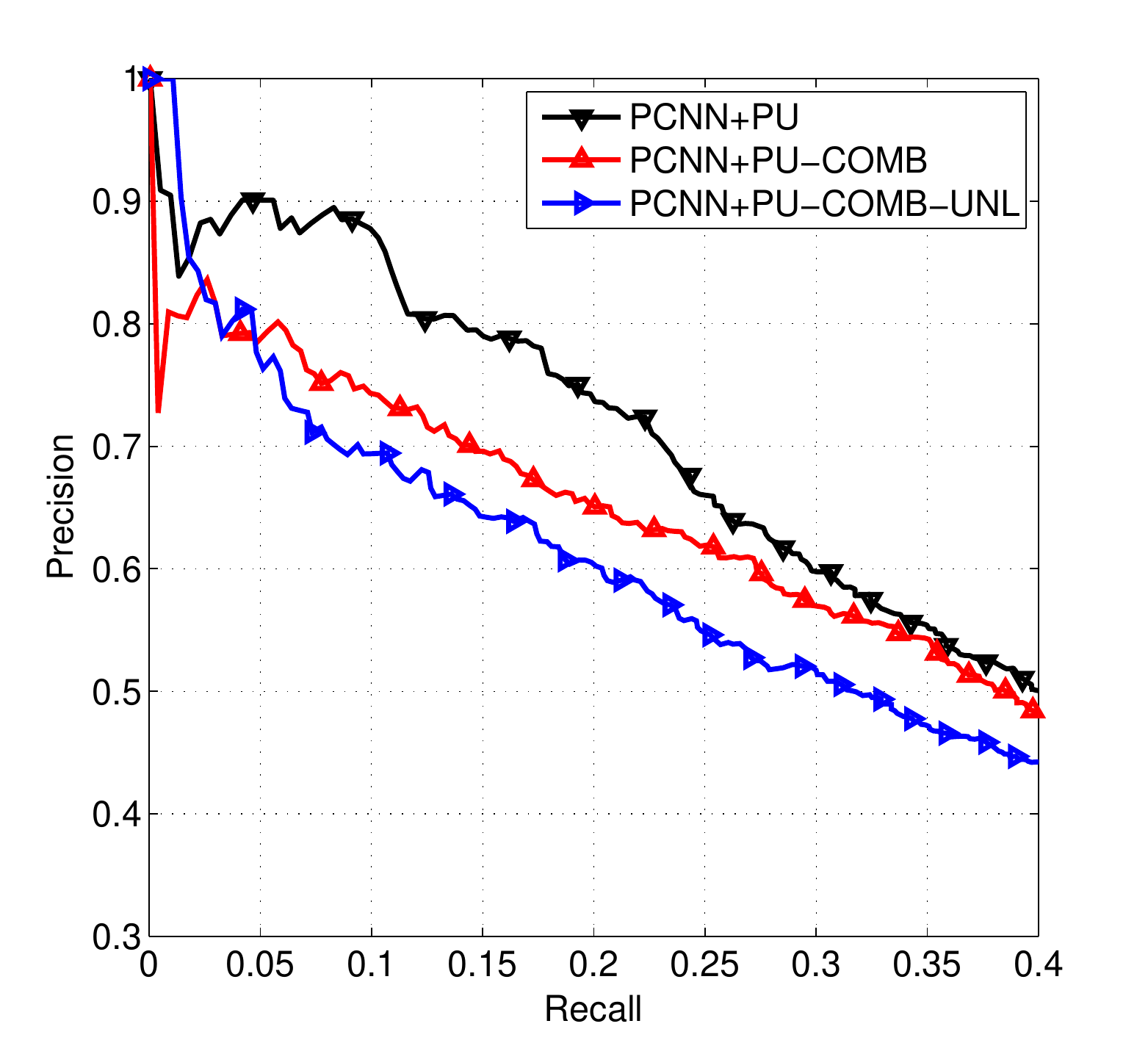}
		\caption{Ablation Study under PCNN}\label{fig:pcnn_ab}
	\end{minipage}
\end{figure*}


Figures \ref{fig:cnn_ab} and \ref{fig:pcnn_ab} show the results. From the figures, we find that if we remove the PU-combined representation, the system performs worse. We also find that the UNL-based representation makes the positive contribution to the systems. These facts indicate that all the three components make positive contributions to our system.
%


\subsection{Effect of Sentence Selector}

We randomly select 300 entity-relation tuples in testing set to evaluate manually, the proportion of correctly labeled is 0.76\footnote{For a given entity-relation tuple, if a sentence in the generated bag can express the corresponding relation, then the sentence is considered to be correctly labeled.}. To demonstrate the precision of the positive sentences selected by our instance selector, we evaluate our proposed sentence selection module by manual method. We report the $P$@$100$ and $P$@$300$ entity-relation tuples generated by the sentence selection module.
The results are shown in Table \ref{table:manual_select}.
From the table,
we can see that
both CNN+PU and PCNN+PU can achieve high accuracy by the sentence selection module. It indicates that the selected sentences by our selector is reliable.

\begin{table*}[tb]
	\centering
	\small
	\begin{tabular}{|p{0.08\textwidth}<{\centering}|  p{0.55\textwidth}   |   p{0.1\textwidth}<{\centering} |   p{0.1\textwidth}<{\centering}   |}
		\hline
		\textbf{Tuple} & \textbf{Generated Sentences via Distant Supervision}\centering & \textbf{capital} & \textbf{contains} \\
		\hline
		\multirow{4}[12]{0.10\textwidth }{\emph{capital } (Liberia, Monrovia)}
		& 1. Nearly two years after Charles Taylor fled [Monrovia] under pressure from advancing rebels and a force of marines on ships off [Liberia], he sits exiled in ... 
		& \multirow{1}[7]{*}{unlabeled} & \multirow{1}[7]{*}{unlabeled} \\
		\cline{2-4}
		
		& 2. On the streets of [Monrovia], [Liberia] 's capital, rumors of a coup or an attack by taylor supporters, several of whom hold ...
		& \multirow{1}[4]{*}{positive} & \multirow{1}[4]{*}{positive}  \\
		\cline{2-4}
		
		& 3. My sister was living in the [Monrovia] suburb of paynesville, [Liberia], with her family and a handful of orphans and other ...
		& \multirow{1}[4]{*}{unlabeled} & \multirow{1}[4]{*}{positive} \\
		\cline{2-4}
		
		& 4. They said Mr. Van Kouwenhoven, who initially ran a car import business, a hotel and a casino in [Monrovia], [Liberia], established ties with ...
		& \multirow{1}[4]{*}{unlabeled} & \multirow{1}[4]{*}{positive} \\
		\hline
		
	\end{tabular}
	
	\caption{Case study: a real example for comparison}\label{case_study}
\end{table*}
\subsection{Case Study}

We further present a real example in Table \ref{case_study} for case study. The entity-relation tuple is
(\emph{Liberia, capital, Monrovia}). There are four sentences containing the entity pair. Here, ``capital" and ``contains" are both relation types defined in Freebase. From the table, we can see that for relation ``capital", the RL agent selects the second sentence as positive example; for relation ``contains", the agent selects three sentences as positive examples and the first sentence as unlabeled example.

\section{Related Work}
Most existing supervised RE systems \cite{riedel2013relation,zhou2016attention} require a great deal of human-annotated relation-specific training data, which is time consuming and labor intensive. To address this issue, distantly supervised RE is proposed to 
automatically generate training data by aligning plain text with KB \cite{mintz2009distant,takamatsu2012reducing,ritter2013modeling}.
To further alleviate the wrong labeling problem arisen by distant supervision assumption,
many studies formulate RE as a MIL problem \cite{riedel2010modeling,hoffmann2011knowledge,surdeanu2012multi}.
\citet{zeng2015distant} further combine at-least-one multi-instance learning and assume that only one sentence expresses the relation for each entity pair.
\citet{Lin2016Neural} propose to use attention mechanism to  dynamically
reduce the weights of  noisy instances, which shows promising results. Besides,  \citet{ji2017distant} attempt to incorporate the external information of entities for RE.
In addition, \citet{he2019AIJ} utilize tree models to use syntax information to enhance the representations of entities.

At present, many researchers have begun to apply deep reinforcement learning  in NLP tasks \cite{narasimhan2016improving,li2016deep}.
%
As for distantly supervised relation extraction, ~\citet{zeng2018rl} directly regard the relation extractor as an RL agent, and predict the bag relation based on the extracted sentence-level relations. 
~\citet{feng2018rl} propose to select correct sentences under the framework of reinforcement learning and then train relation extractor on the cleaned data.
\citet{Qin2018robust} also use RL to determine whether a sentence is positive or negative and both positive and negative samples are used to build a classifier for rewards.
Differently from the one-agent-for-one-relation strategy of  \citet{Qin2018robust}, we build a single agent for all relation types.
In addition, in  existing approaches,  they only use the selected positive instances and ignore the other sentences when making relation prediction at bag-level, or directly
treat the unlabeled sentences which might contain positive instances as negative instances. In contrast, in our approach, all the unlabeled instances directly contribute to our final relation extractor.


\section{Conclusion}

In this paper, we propose a novel approach for bag-level relation classification which makes full use of the results of the sentence selector. In the first step in our approach, we apply the idea of reinforcement learning to build the sentence selector which selects positive and unlabeled sentences from noisy data. In the sentence selector, we newly add the information of relation embedding in the state representation. Then, when training the bag-level relation classifier, we consider the unlabeled instances which are often discarded by the previous studies. With proper joint training, sentence selector and relation classifier interact with each other. Finally, the experimental results show that our proposed system consistently outperforms the baseline systems. Further studies indicate that adding relation embedding in the state representation and using unlabeled instances for training models are helpful to relation extraction.

\section{ Acknowledgments}
The research work is supported by the National Key Research and Development Program of China (Grant No. 2017YFB1002104) and the National Natural Science Foundation of China (Grant Nos. 61876115, 61936010). This work is
partially supported by the joint research project of Alibaba and Soochow
University and a Project Funded by the Priority Academic Program Development of Jiangsu Higher Education Institutions (PAPD). Corresponding author is Wenliang Chen. We would also thank the anonymous reviewers for their detailed comments, which have helped us to improve the quality of this work.

\bibliography{emnlp-ijcnlp-2019}
\bibliographystyle{aaai}

\end{document}